\pdfoutput=1

\documentclass[11pt]{article}

\usepackage[]{ACL2023}

\usepackage{times}
\usepackage{siunitx}
\usepackage{latexsym}
\usepackage{booktabs}
\usepackage{multirow}
\usepackage{amsmath}
\usepackage{rotating}
\usepackage{tabularx}
\usepackage{subcaption} 
\usepackage{listings}
\usepackage{enumitem}
\usepackage{xcolor}

\lstdefinestyle{json}{
    basicstyle=\ttfamily\footnotesize,
    breaklines=true,
    frame=single,
    captionpos=b,
    showstringspaces=false,
    escapeinside={(*@}{@*)}
}

\definecolor{blue-multi}{HTML}{1f77b4}
\definecolor{brown-multi}{HTML}{ff7f0e}
\definecolor{green-multi}{HTML}{2ca02c}
\newcommand{\Hquad}{\hspace{0.37em}}

\usepackage[T1]{fontenc}

\usepackage[utf8]{inputenc}

\usepackage{microtype}

\usepackage{inconsolata}

\newcommand{\algoname}{IteRABRe}
\title{\algoname{}: \underline{Ite}rative \underline{R}ecovery-\underline{A}ided \underline{B}lock \underline{Re}duction} 

\author{
Haryo Akbarianto Wibowo$^{1,2}$, Haiyue Song$^1$, \\
\textbf{Hideki Tanaka$^1$, Masao Utiyama$^1$, Alham Fikri Aji$^2$, Raj Dabre$^1$} \\
$^{1}$NICT$\Hquad$ $^{2}$MBZUAI \\
\texttt{\small \{haryo.wibowo,alham.fikri\}@mbzuai.ac.ae, \{haiyue.song,hideki.tanaka,mutiyama,raj.dabre\}@nict.go.jp} \\
}

\begin{document}
\maketitle
\begin{abstract}
Large Language Models (LLMs) have grown increasingly expensive to deploy, driving the need for effective model compression techniques. While block pruning offers a straightforward approach to reducing model size, existing methods often struggle to maintain performance or require substantial computational resources for recovery. We present \algoname{}, a simple yet effective iterative pruning method that achieves superior compression results while requiring minimal computational resources. Using only 2.5M tokens for recovery, our method outperforms baseline approaches by ~3\% on average when compressing the Llama3.1-8B and Qwen2.5-7B models. \algoname{} demonstrates particular strength in the preservation of linguistic capabilities, showing an improvement 5\% over the baselines in language-related tasks. Our analysis reveals distinct pruning characteristics between these models, while also demonstrating preservation of multilingual capabilities.

\end{abstract}

\section{Introduction}

We are in the era of the burgeoning of producing Large Language Models (LLMs), which has led to the necessity of making them smaller due to deployment costs. Several approaches focus on model reduction, such as model sparsification, reducing LLM hidden sizes, or removing presumably unimportant blocks. However, preserving performance while compressing models remains challenging.

Block pruning is a straightforward compression approach for reducing LLM size, motivated by the layer redundancy found in LLM architectures~\cite{men2024shortgptlayerslargelanguage,dumitru-etal-2024-change,chen2025streamliningredundantlayerscompress}. While detecting redundant or unimportant blocks can minimize performance degradation from pruning, some loss is inevitable. Although post-finetuning can help recover performance, simultaneous pruning of multiple blocks may still cause unrecoverable damage.

\begin{figure}
    \centering
    \includegraphics[width=\linewidth]{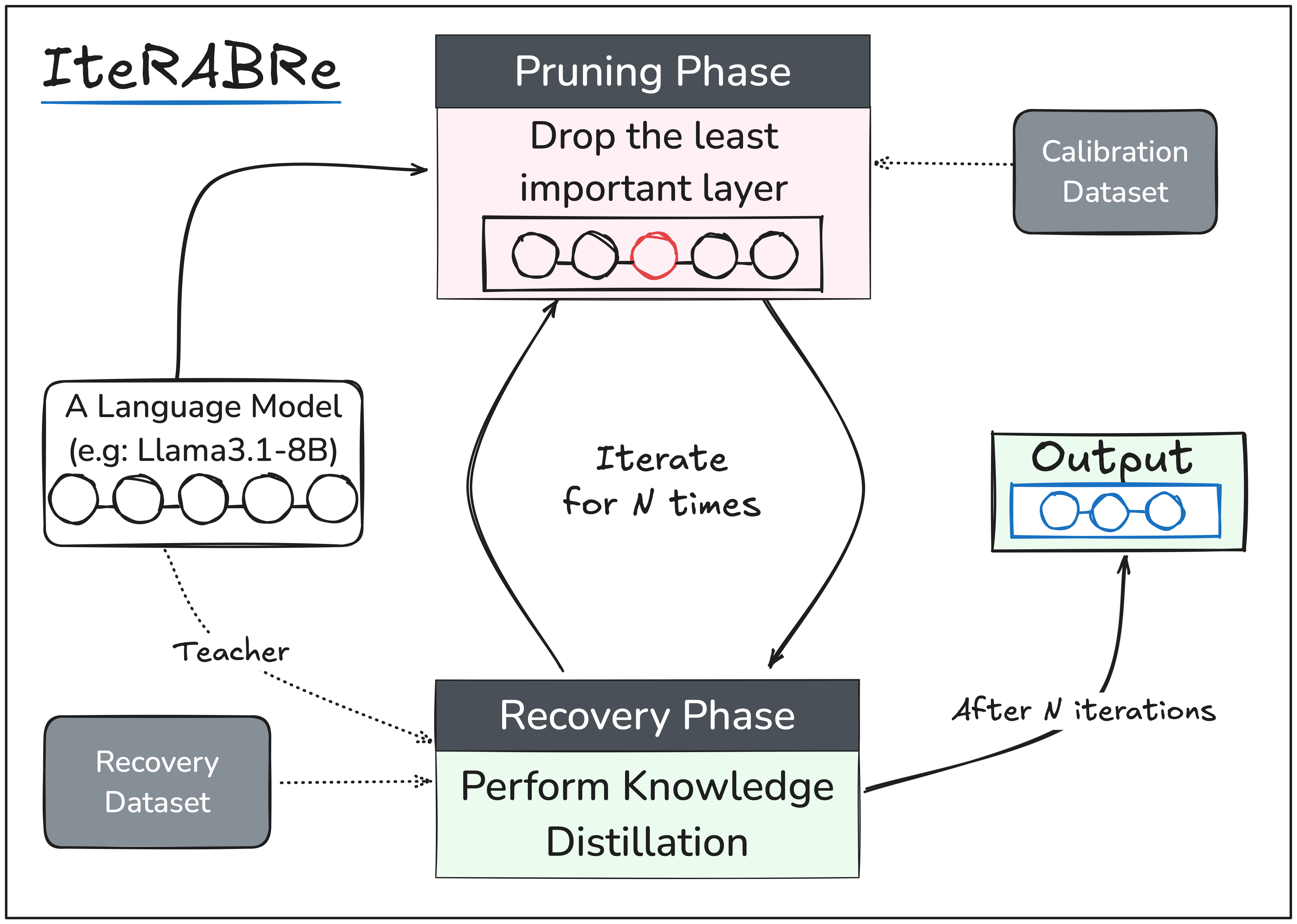}
    \caption{Overall methodology of \algoname{}. It consists of pruning and recovery phase that are done iteratively until the desired compression rate or minimum number of layer achieved.}
    \label{fig: overall}
\end{figure}

One possible solution is to take an iterative approach. \citealp{muralidharan2024compactlanguagemodelspruning} proposes iterative pruning with knowledge distillation to recover from performance loss. This work successfully compresses a 15B LLM into smaller 8B and 4B versions, achieving competitive results compared to other LLMs of similar size. However, this approach may be impractical for those with limited computational resources, as it employs a compute-dependent method in the pruning process and requires 8T tokens for the recovery process.

This leads us to ask: \textit{\textbf{Can we develop an exhustive, efficient and effective iterative block pruning method for model compression?}} We investigate this question by introducing \algoname{}, a straightforward iterative pruning approach. We choose layer pruning for its simplicity and enhanced interpretability in preservation. To test efficiency, we perform recovery using only 2.5M tokens of data. Our method outperforms other baselines by approximately 3\% on average for Llama3.1-8B and Qwen2.5-7B models. Furthermore, our approach shows particular strength in preserving linguistic tasks, demonstrating 5\% better performance than baselines. Additionally, this approach also exhibits zero-shot cross-lingual capabilities, such as retaining the German language using solely English data recovery. Our key contributions are:
\begin{enumerate}
\item We introduce \algoname{}, a simple and efficient iterative pruning approach with recovery.
\item We provide detailed analysis of performance degradation, including recovery phase impact and patterns in layer index drop sequences, probing different behavior exhibited in each model family. 
\item We present granular analysis of \algoname{}'s impact on knowledge, linguistic, and multilingual zero-shot capabilities.
\end{enumerate}

\section{Background: Structured Prunning} \label{sec: background}
 
The increasing scale of Large Language Models (LLMs) has driven efforts to reduce their computational and memory footprint for deployment. A prominent approach is \textbf{structured pruning}~\cite{wang-etal-2020-structured}, where some components (e.g., layers, attention heads) are removed from a large model \( M_L \) to derive a smaller model \( M_S \). However, pruning often causes performance degradation, making a careful selection of components and recovery strategies after pruning necessary~\cite{sun2024simpleeffectivepruningapproach, yin2024outlierweighedlayerwisesparsity,ma2023llmprunerstructuralpruninglarge}. The following are the explanations of these phases:

\paragraph{Pruning Phase} Formally, let \( M_L \) consist of \( N \) transformer component blocks \( \{B_1, B_2, ..., B_N\} \). Pruning involves ranking blocks by importance and retaining the top-\( k \) blocks (\( k < N \)) to form \( M_S \). The importance of a block \( B_i \) is determined by a scoring function \( f(B_i) \), which can be defined as: 
\[
f(B_i) = \text{Importance}(B_i; \mathcal{D}_{\text{eval}})
\]
Here, \( \mathcal{D}_{\text{eval}} \) is a validation dataset (calibration dataset) used to compute metrics to determine the blocks' importance. Blocks are then sorted by \( f(B_i) \), and the least important \( N - k \) are pruned or dropped: 
\[
M_S = \text{Prune}(M_L, k)
\]
\paragraph{Recovery Phase} To alleviate performance degradation due to pruning, this phase fine-tunes \( M_S \) on a recovery dataset \( \mathcal{D}_r \) on the respective tasks, such as Causal Language Modeling.  The recovery process is useful to adapt to its new structure and reallocate its internal knowledge to its remaining capacity.

The recovery process optimizes:
\[
\theta_S^* = \arg\min_{\theta_S} \mathcal{L}(M_S(\theta_S; \mathcal{D}_r), y)
\]
where \( \theta_S \), \( y \) denotes the parameters of \( M_S \) and ground truth, respectively.\footnote{These variables declared in this section will be used throughout this paper.}

\section{Proposed Method: \algoname}
We introduce \algoname{}, an iterative compression framework for large language models that alternates between pruning and recovery phases until a target model size, which is the number of layers, is reached. This process continues until the desired number of layers in $M_s$. Although iterative compression has been explored previously~\cite{muralidharan2024compactlanguagemodelspruning}, our approach makes two key distinctions: (1) a direct layerwise pruning strategy and (2) an efficient recovery process that achieves competitive performance with significantly reduced data requirements. The method introduced in \citealp{muralidharan2024compactlanguagemodelspruning} may achieve better results through extensive compute and data resources, however our approach demonstrates competitive performance to other baselines with substantially lower resource requirements.  Our approach is illustrated in Figure~\ref{fig: overall}.

The following subsections elaborate on the respective phases in \algoname{} defined in \S\ref{sec: background}.

\subsection{\algoname's Pruning Phase}

We define $B_i$ as transformer blocks, where each block consists of \textbf{self-attention and feed-forward components}. To minimize performance degradation during pruning, we evaluate the importance of each layer $B_i$ by measuring its contribution to the model's output quality. 

Specifically, we compute the cosine similarity between the last hidden state of the original model $M_L$ and the last hidden state of the candidate pruned models $M_{cs}^{(i)}$, where $M_{cs}^{(i)}$ is obtained by removing one layer of self-attention from $M_L$.
The importance score $f(B_i)$ for a block $B_i$ is defined as:
\[
f(B_i) =\frac{1}{|\mathcal{D}_\text{eval}|} \sum_{d=1}^{|\mathcal{D}_\text{eval}|}\text{sim}\left(h(M_L)_d, h(M_{cs}^{(i)})_d\right)
\]
where $h(M_L)_d$ is the last hidden state of the original model $M_L$ with $N$ layers for the $d$-th input sequence in $\mathcal{D}_\text{eval}$, $ h(M_{cs}^{(i)})_d$ is the last hidden state of the pruned model $M_{cs}^{(i)}$ with $N-1$ layers for the same input sequence. $\text{sim}(\cdot, \cdot)$ denotes the cosine similarity function.

After computing $f(B_i)$ for all blocks, we sort the blocks by their similarity scores. The lowest similarity block will be selected for removal, as it indicates the least impact on model performance. This process yields our final pruned model $M_{cs}$ with the selected blocks removed. $M_{cs}$, then will be processed in the Recovery Phase. 

For better clarity in the following sections, we also denote $M_{cs}^{[j]}$ as the final pruned model chosen in iteration $j$.

\begin{table*}[t]

\centering
\setlength{\tabcolsep}{3.5pt}

\resizebox{1\linewidth}{!}{
\begin{tabular}{@{}llcc*{10}{c}r@{}}
\toprule
\multirow{2}{*}{Model} & \multirow{2}{*}{Approach} & \multirow{2}{*}{\#L} & \multirow{2}{*}{Wiki$\downarrow$} & \multicolumn{5}{c}{Reasoning} & \multicolumn{3}{c}{Language Comprehension} & \multicolumn{2}{c}{Knowledge} \\
\cmidrule(lr){5-9} \cmidrule(lr){10-12} \cmidrule(l){13-14}
 & & & & ARC-C & ARC-E & HellaSwag & COPA & PIQA & BLiMP & RACE & Winogrande & BoolQ & MMLU \\
\midrule
\multirow{4}{*}{Llama3.1 8B} 
 & Not Pruned & 32 & 8.65 & 51.28 & 81.48 & 60.03 & 87.0 & 80.14 & 81.93 & 39.14 & 73.56 & 82.08 & 63.59 \\
 & LaCO & 24 & 23.55 & 30.29 & 63.01 & 43.22 & \textbf{81.0} & 71.76 & 79.34 & 30.91 & 55.72 & 61.99 & 23.96 \\
 & ShortGPT & 24 & 6636.72 & 27.47 & 42.68 & 28.28 & 63.0 & 60.55 & 66.84 & 25.07 & 53.91 & 37.58 & \textbf{32.21} \\
 & \algoname & 24 & \textbf{16.89} & \textbf{33.02} & \textbf{67.85} & \textbf{47.49} & 80.0 & \textbf{74.27} & \textbf{84.10} & \textbf{35.69} & \textbf{60.93} & \textbf{62.26} & 23.80 \\
\cmidrule[0.5pt](lr){2-14}
\multirow{4}{*}{Llama3.2 3B}
 & Not Pruned & 28 & 11.06 & 42.24 & 74.54 & 55.27 & 86.0 & 76.61 & 82.15 & 40.19 & 69.77 & 73.24 & 54.38 \\
 & LaCO & \textbf{24*} & \textbf{25.55} & 26.96 & 56.65 & \textbf{42.54} & \textbf{80.0} & \textbf{71.76} & 80.15 & \textbf{32.63} & 55.72 & 60.03 & 24.47 \\
 & ShortGPT & 21 & 235.23 & \textbf{30.89} & 49.71 & 37.34 & 71.0 & 64.15 & 72.53 & 30.53 & \textbf{61.88} & 45.02 & \textbf{34.38} \\
 & \algoname & 21 & 26.92 & 30.12 & \textbf{58.84} & 41.53 & 76.0 & 70.08 & \textbf{80.60} & 32.44 & 58.33 & \textbf{62.17} & 26.15 \\
\cmidrule[0.5pt](lr){2-14}
\multirow{4}{*}{Llama3.2 1B}
 & Not Pruned & 16 & 13.91 & 31.31 & 65.40 & 47.78 & 77.0 & 74.48 & 82.44 & 37.89 & 60.77 & 63.98 & 37.54 \\
 & LaCO & 12 & 80.16 & 19.28 & 40.15 & 29.77 & 56.0 & 61.10 & 72.93 & 26.03 & 51.38 & 37.83 & \textbf{23.05} \\
 & ShortGPT & 12 & 846.24 & \textbf{25.51} & 34.89 & 31.51 & \textbf{70.0} & 59.03 & 74.20 & 24.88 & \textbf{54.14} & 55.96 & 22.70 \\
 & \algoname & 12 & \textbf{31.85} & 23.89 & \textbf{51.14} & \textbf{33.39} & 66.0 & \textbf{65.61} & \textbf{82.71} & \textbf{28.04} & 51.22 & \textbf{62.14} & 23.00 \\
\cmidrule[0.5pt](lr){2-14}
\multirow{4}{*}{Qwen2.5-7B}
 & Not Pruned & 28 & 10.35 & 47.78 & 80.39 & 60.03 & 91.0 & 78.67 & 82.24 & 41.63 & 72.93 & 84.65 & 71.91 \\
 & LaCO & \textbf{22*} & 48.38 & 29.52 & 50.80 & 39.32 & 71.0 & 67.14 & 75.60 & 27.18 & 55.88 & 47.19 & \textbf{31.83} \\
 & ShortGPT & 21 & 18.57 & 33.79 & 70.88 & 44.32 & 76.0 & 74.27 & 81.93 & 33.01 & 53.51 & 45.84 & 26.52 \\
 & \algoname{} & 21 & \textbf{16.40} & \textbf{35.58} & \textbf{71.13} & \textbf{45.59} & \textbf{77.0} & \textbf{74.32} & \textbf{83.48} & \textbf{36.08} & \textbf{57.70} & \textbf{53.73} & 30.94 \\
\cmidrule[0.5pt](lr){2-14}
\multirow{4}{*}{Qwen2.5-0.5B}
 & Not Pruned & 24 & 21.68 & 29.52 & 64.56 & 40.64 & 74.0 & 70.29 & 81.73 & 35.02 & 56.35 & 62.42 & 47.73 \\
 & LaCO & 18 & 230.05 & \textbf{24.06} & 45.79 & 30.95 & 60.0 & 62.51 & 71.20 & 26.41 & \textbf{51.22} & 54.13 & \textbf{25.54} \\
 & ShortGPT & 18 & 45.86 & 21.42 & 52.15 & 32.39 & 62.0 & 65.18 & 77.94 & \textbf{28.80} & 49.33 & \textbf{58.96} & 25.17 \\
 & \algoname & 18 & \textbf{38.17} & 23.89 & \textbf{54.71} & \textbf{32.41} & \textbf{68.0} & \textbf{65.40} & \textbf{78.61} & 27.27 & 50.75 & 47.16 & 23.32 \\
\bottomrule
\end{tabular}
}
\caption{Performance comparison across model scales and tasks, showing perplexity (Wiki$\downarrow$, where lower is better) and accuracy scores (\%). Bold indicates the best performance among other approaches (LaCO, ShortGPT, \algoname{}) for each metric. *: Due to the dependency on hyperparameter in  LaCO, some of its results may have incomparable compression with others. \#L denotes number of layers.}
\label{tab:main-results}
\end{table*}

\subsection{\algoname's Recovery Phase}\label{section:recovery-phase}

To further preserve the degradation quality of the model, we employ knowledge distillation, where we put the original model, $M_L$ as the teacher $T$ and the pruned model from the previous phase in the same iteration $j$ as its student $M_{cs}^{[j]}$, which we denote here as $S$. We follow the TinyBERT design~\cite{jiao-etal-2020-tinybert}, where we compute the mean square error (MSE) between all hidden states, attention, and output logits. We use MSE for the output logits as it shows better performance than KL Divergence~\cite{kim2021comparingkullbackleiblerdivergencemean}. Formally, it is defined as follows:
$$
\begin{aligned}
\mathcal{L}_{KD} =  \sum_{l=1}^{L}  &\left( \text{MSE}(\mathbf{H}_T^{\text{map}(l)}, \mathbf{H}_S^{l}) + \right. \\
& \left.  \text{MSE}(\mathbf{A}_T^{\text{map}(l)}, \mathbf{A}_S^{l}) \right ) + \\
& \text{MSE}(\mathbf{z}_T, \mathbf{z}_S)
\end{aligned}
$$
Here, $\mathbf{H}_{T}^{map(l)}$ and $\mathbf{H}_S^l$ represent the hidden states in layers $l$ and $map(l)$ for the teacher and student models, respectively, while $\mathbf{A}_T^{map(l)}$ and $\mathbf{A}_S^l$ denote their corresponding attention matrices. The output logits of the teacher and student models are represented by $\mathbf{z}_T$ and $\mathbf{z}_S$, respectively.  $map(l)$ is defined as the mapping of a student's layer to the teacher's layer which aligns the student's layer index \( l \) with the corresponding original index in the teacher model \footnote{See Appendix~\ref{sec:apx-mapping} for more explanation}.

After this phase, we produce a recovered pruned model  $M_{cs-rec}^{[j]}$ as the final  chosen in iteration $j$. $M_{cs-rec}^{[j]}$ is then processed to the next iteration $j+1$

\begin{figure*}
    \centering

    \begin{subfigure}{0.49\textwidth}
        \centering
        \includegraphics[width=\textwidth]{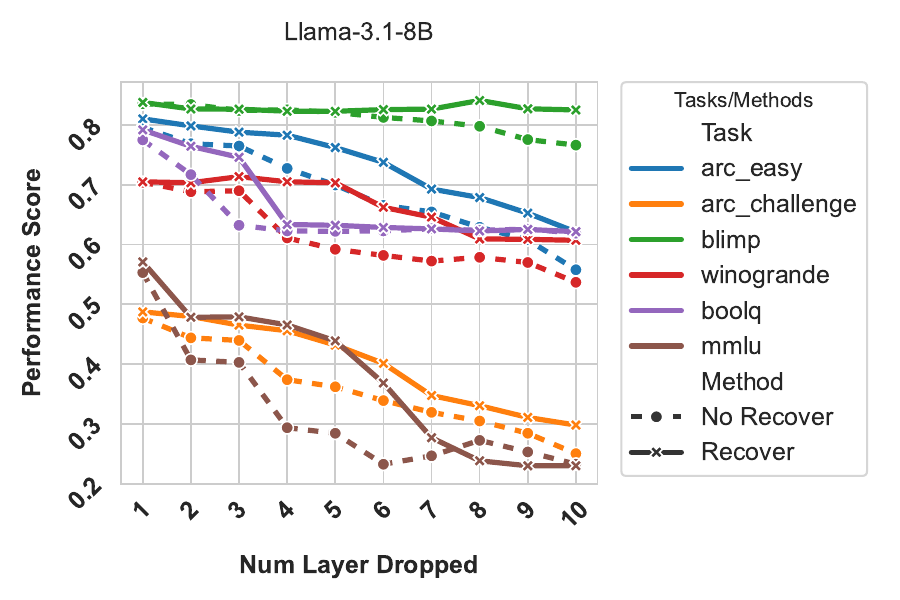}
    \end{subfigure}
    \hfill
    \begin{subfigure}{0.49\textwidth}
        \centering
        \includegraphics[width=\textwidth]{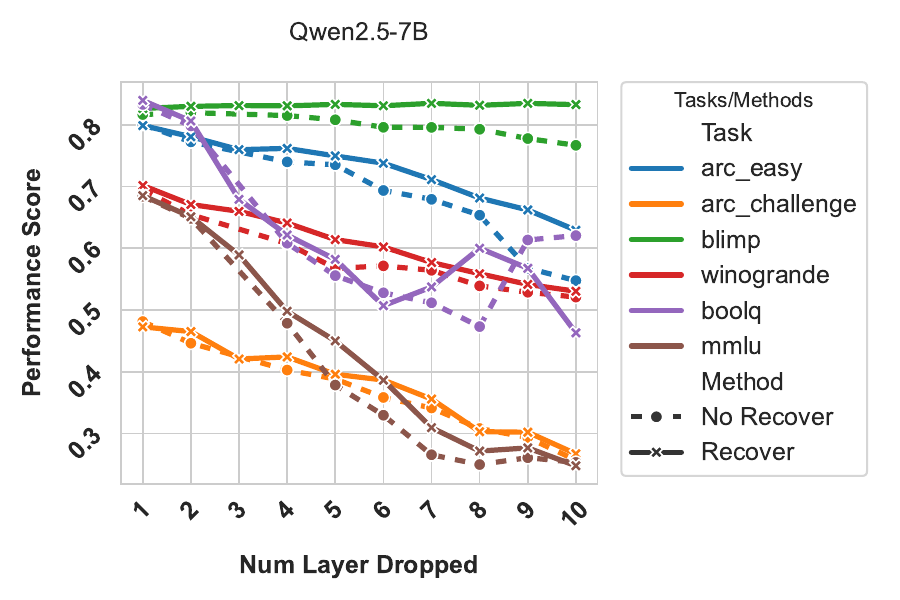}
    \end{subfigure}
    \vspace{-1.2em}
    \begin{subfigure}{0.49\textwidth} 
        \centering
        \includegraphics[width=\textwidth]{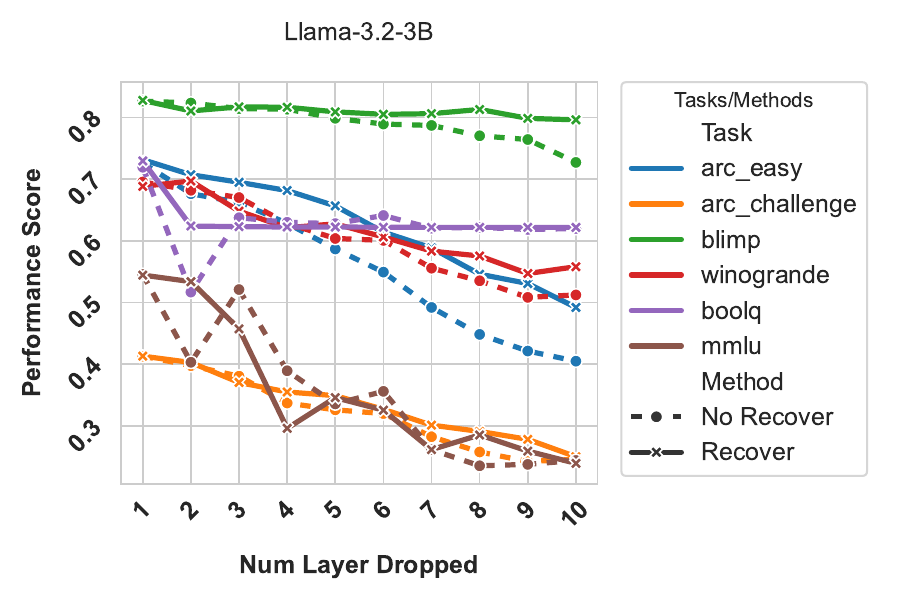}
    \end{subfigure}
    \begin{subfigure}{0.49\textwidth}
        \centering
        \includegraphics[width=\textwidth]{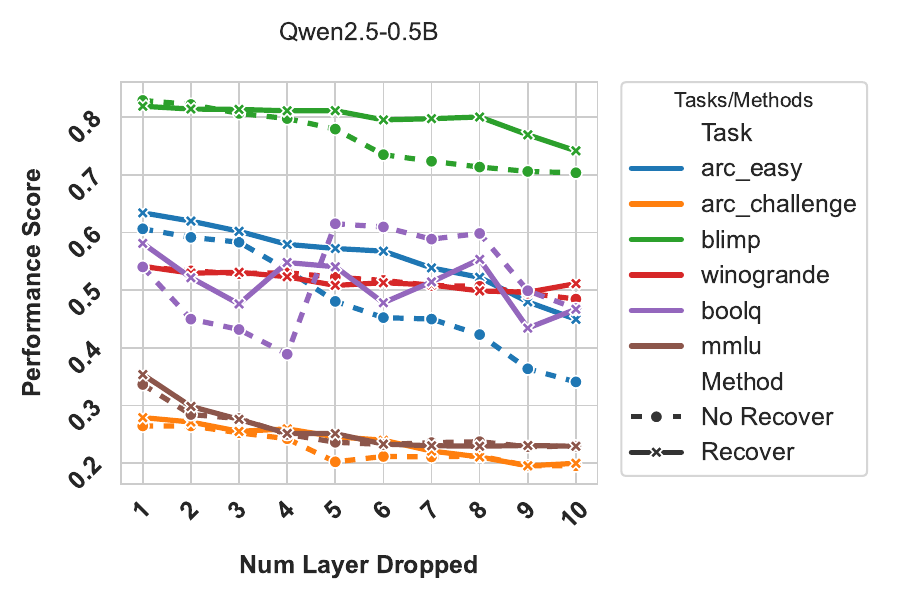}
    \end{subfigure}
    \caption{\algoname{}'s performance on six different subtasks. dotted line denoted implementing \algoname{} without recovery phase while solid line denoted layer prunning and recovery phase are done in \algoname{}}
    \label{fig:recovvsnorecov}
\end{figure*}

\section{Experimental Setup}

To test our proposed algorithm and analyze it, we use the following setups:

\paragraph{Evaluation} To have a better assessment of our approach, we categorize the benchmark dataset into three categories, \textbf{reasoning, language comprehension, and knowledge}. For reasoning, we leverage \texttt{arc-challenge}  and \texttt{arc-easy dataset}~\cite{clark2018thinksolvedquestionanswering}, \texttt{hellaswag}~\cite{zellers2019hellaswagmachinereallyfinish}, 
\texttt{COPA}~\cite{roemmele2011choice},  \texttt{PIQA}~\cite{bisk2020piqa}. For language comprehension, we test our data on \texttt{BLiMP}~\cite{warstadt2020blimp}, \texttt{RACE}~\cite{lai-etal-2017-race}, and \texttt{Winogrande}~\cite{sakaguchi2021winogrande}. As for the knowledge category, we use \texttt{BoolQ}~\cite{clark2019boolq} and \texttt{MMLU}~\cite{hendryckstest2021}.  To evaluate our model, we use off-the-shelf \texttt{lm-eval-harness}~\cite{eval-harness} library. We use the context length of 1024 and employ the zero-shot setting to obtain the score. To measure the performance, we use test set of \texttt{wikitext-2-raw-v1} to measure perplexity and use accuracy for the rest. 

\paragraph{Models} We used two widely used LLM families, Qwen2.5~\cite{qwen2.5} and Llama3~\cite{llama3}. To observe the impact of model size, we use 8B, 3B, and 1B from Llama3 and 7B and 0.5B for Qwen2.5.

\paragraph{Pruning Phase} For the pruning phase, we use 10 instances as the calibration dataset, sampled from the \texttt{wikitext-2-raw-v1} dataset on Hugging Face, following Yang et al. (2024). The sampled Wikitext data is provided in Appendix 
\ref{sec: apx-calibration}. As demonstrated by \citet{yang-etal-2024-laco}, using different samples does not significantly affect the results.

\paragraph{Recovery Phase} We employ Knowledge Distillation~\cite{hinton2015distillingknowledgeneuralnetwork} following the TinyBERT approach~\cite{jiao-etal-2020-tinybert}. To accommodate our computational constraints, we implement LoRA~\cite{hu2021loralowrankadaptationlarge} with a rank of 32. Our training configuration includes a batch size of 4 with gradient accumulation of 8 (effective batch size of 32), learning rate of $1\times10^{-4}$, and maximum sequence length of 512. For efficient recovery training, we conduct a single epoch on the \texttt{Wikitext-2-raw-v1} dataset, comprising approximately 2.5M tokens. The training was performed on 2 $\times$ A100 GPUs.

\begin{table*}[t]
\scriptsize
\centering
\setlength{\tabcolsep}{2pt}  
\begin{tabular}{ccc|cccccccc|cc||c}
\toprule
\textbf{Approach} & \textbf{\#Dropped Layers} & \textbf{Wiki} & \textbf{BOOLQ} & \textbf{ARC-E} & \textbf{COPA} & \textbf{BLiMP} & \textbf{H-SWAG} & \textbf{PIQA} & \textbf{ARC-C} & \textbf{RACE} & \textbf{WG} & \textbf{MMLU} & \textbf{Approach Avg Diff} \\
\midrule
\multicolumn{14}{c}{\textbf{Llama 3.1-8B}} \\
\midrule
Direct+R & 4 & 12.22 & 65.05 & 76.60 & 87.00 & 82.98 & 54.69 & 77.04 & 44.54 & 39.23 & 71.03 & 43.31 & \multirow{3}{*}{\textbf{0.15}} \\
Iterative+R & 4 & 12.12 & 63.33 & 78.28 & 85.00 & 82.30 & 55.18 & 77.15 & 45.56 & 39.14 & 70.48 & 46.53 & \\
\textit{Diff} & & \textit{0.10} & \textit{-1.72} & \textit{1.68} & \textit{-2.00} & \textit{-0.68} & \textit{0.49} & \textit{0.11} & \textit{1.02} & \textit{-0.09} & \textit{-0.55} & \textit{3.22} & \\
\midrule
Direct+R & 8 & 20.64 & 51.52 & 66.37 & 79.00 & 81.85 & 46.49 & 72.31 & 33.53 & 35.79 & 66.30 & 37.52 & \multirow{3}{*}{\textcolor{red}{-0.10}}\\
Iterative+R & 8 & 16.89 & 62.26 & 67.85 & 80.00 & 84.10 & 47.49 & 74.27 & 33.02 & 35.69 & 60.93 & 23.80 & \\
\textit{Diff} & & \textit{3.75} & \textit{11.01} & \textit{1.48} & \textit{1.00} & \textit{2.25} & \textit{1.00} & \textit{1.96} & \textit{-0.51} & \textit{-0.10} & \textit{-5.37} & \textit{-13.72} & \\
\midrule
Direct+R & 12 & 35.74 & 38.13 & 55.43 & 70.00 & 81.32 & 39.32 & 67.79 & 27.73 & 32.92 & 57.06 & 24.67 & \multirow{3}{*}{\textbf{2.61}} \\
Iterative+R & 12 & 31.63 & 62.17 & 58.04 & 75.00 & 82.29 & 38.56 & 67.36 & 27.22 & 32.06 & 54.78 & 22.96 & \\
\textit{Diff} & & \textit{4.11} & \textit{24.04} & \textit{2.61} & \textit{5.00} & \textit{0.97} & \textit{-0.76} & \textit{-0.44} & \textit{-0.51} & \textit{-0.86} & \textit{-2.29} & \textit{-1.71} & \\
\midrule
\multicolumn{14}{c}{\textbf{QWEN2.5-7B}} \\
\midrule
Direct+R & 4 & 12.99 & 54.95 & 76.01 & 83.00 & 83.61 & 51.82 & 77.58 & 41.72 & 36.27 & 64.88 & 53.14 & \multirow{3}{*}{\textbf{0.77}} \\
Iterative+R & 4 & 13.23 & 62.14 & 76.18 & 86.00 & 83.07 & 52.43 & 77.80 & 42.41 & 36.84 & 64.09 & 49.79 & \\
\textit{Diff} & & \textit{-0.24} & \textit{7.19} & \textit{0.17} & \textit{3.00} & \textit{-0.54} & \textit{0.61} & \textit{0.22} & \textit{0.68} & \textit{0.57} & \textit{-0.79} & \textit{-3.35} & \\
\midrule
Direct+R & 8 & 18.96 & 48.20 & 67.89 & 78.00 & 83.34 & 42.43 & 72.47 & 32.08 & 32.73 & 54.70 & 26.17 & \multirow{3}{*}{\textbf{1.12}} \\
Iterative+R & 8 & 18.79 & 60.00 & 68.14 & 75.00 & 83.16 & 42.43 & 73.23 & 30.29 & 33.88 & 55.88 & 27.15 & \\
\textit{Diff} & & \textit{0.17} & \textit{11.80} & \textit{0.25} & \textit{-3.00} & \textit{-0.18} & \textit{0.00} & \textit{0.76} & \textit{-1.79} & \textit{1.15} & \textit{1.18} & \textit{0.98} & \\
\midrule
Direct+R & 12 & 46.15 & 57.09 & 53.75 & 63.00 & 78.97 & 33.38 & 66.05 & 22.95 & 29.00 & 52.88 & 24.38 & \multirow{3}{*}{\textcolor{red}{-0.20}}\\
Iterative+R & 12 & 35.22 & 42.69 & 57.15 & 66.00 & 81.47 & 35.50 & 66.05 & 24.66 & 28.71 & 53.04 & 24.18 & \\
\textit{Diff} & & \textit{10.93} & \textit{-14.40} & \textit{3.41} & \textit{3.00} & \textit{2.50} & \textit{2.12} & \textit{0.00} & \textit{1.71} & \textit{-0.29} & \textit{0.16} & \textit{-0.20} & \\
\midrule
\midrule
Tasks Avg Diff & - & \textbf{3.13} & \textbf{6.32} & \textbf{1.60} & \textbf{1.17} & \textbf{0.72} & \textbf{0.57} & \textbf{0.44} & \textbf{0.10} & \textbf{0.06} & \textcolor{red}{-1.28} & \textcolor{red}{-2.46} & - \\
\bottomrule
\end{tabular}
\caption{The comparison between direct and iterative approach. Diff denote the difference between direct with respect to iterative approach.}
\label{tab:direct_onbyone}
\end{table*}

\paragraph{Baseline} To evaluate \algoname{}, we compare it with two baseline layer pruning methods: LaCO~\cite{yang-etal-2024-laco} and ShortGPT~\cite{men2024shortgptlayerslargelanguage}. While our approach adopts LaCO's layer importance assessment methodology, ShortGPT employs Block Influence (BI). Our method extends these approaches by incorporating recovery and iterative pruning. We target a compression rate of approximately 25\%, following previous works. For ShortGPT, we implemented the method ourselves to obtain results, while for LaCO, we utilized their publicly available code. Since LaCO's compression rate varies with hyperparameters, we conducted a grid search and selected the model with the closest compression rate and highest perplexity score on \texttt{wikitext-v2-raw-v1}.

\section{Results} \label{sec:exp-results}

\paragraph{\algoname{} Outperforms Other Baselines Overall} Table~\ref{tab:main-results} presents the experimental results. \algoname{} outperforms other methods (LaCO and ShortGPT) across all model scales.  Specifically, \algoname{} maintains a lower perplexity on Wikitext compared to the baselines, avoiding the sharp increases observed with ShortGPT on Llama3.1-8B (6636.72) and LaCO on Qwen2.5-7B (48.38).  \algoname{} also achieves the highest performance in the reasoning domain for the 7B and 8B models. However, for smaller models (0.5B, 1B, and 3B), \algoname{} exhibits a small performance gap compared to the baselines on \texttt{arc-challenge}, likely because this task's reliance on multi-hop reasoning demands greater model capacity.

In the language category, \algoname{} maintains performance better than the other methods, particularly on \texttt{BLIMP}, where the large models (7B and 8B) even outperform their non-pruned counterparts. We attribute this to the recovery phase, where training on \texttt{wikitext} helps preserve linguistic capabilities.  On the other hand, \texttt{RACE} and \texttt{Winogrande} show moderate performance gaps (2-5\%) across all model sizes. These results suggest that our method offers particular advantages for language comprehension in large models.

In the knowledge domain, \algoname{} achieves strong \texttt{BoolQ} performance, with the exception of Qwen2.5-0.5B.  The improved accuracy for this model is likely due to the use of \texttt{wikitext} as a recovery training dataset. However, MMLU results lag behind the other methods, by approximately 9\% compared to the highest performer on Llama3.1-8B and Llama3.2-3B, and by 1-2\% on the other models. This difference may be due to the fact that our approach does not preserve or recover information crucial for maintaining MMLU performance.

\paragraph{\algoname{}'s recovery phase boosts performance, notably for larger models on reasoning and language tasks.} We investigated the impact of each phase of \algoname{}. The results are shown in Figure~\ref{fig:recovvsnorecov}.  In summary, the iterative recovery phase helps preserve performance on reasoning and language tasks, particularly in later iterations. For example, with Llama3.1-8B, the performance difference between the first and third iterations is approximately 1-3\%, while it widens to 5-10\% between the fourth and sixth iterations. This pattern is also observed on Winogrande. For BLIMP, the performance gap similarly increases in later iterations (6th-10th). QWEN exhibits the same trend, albeit with smaller gaps.

For knowledge tasks, MMLU shows a clear performance difference in both the 7B and 8B models. However, BoolQ exhibits an irregular trend with Qwen2.5-7B, with fluctuating performance (sometimes higher, sometimes lower) and \textasciitilde1\% differences in the Llama3 model. This behavior is also observed in smaller models (0.5B and 3B) for both tasks. Overall, the recovery phase provides a considerable performance improvement, except in the knowledge domain, especially for smaller models.

\paragraph{Iterative pruning generally outperforms direct pruning on most tasks, but its effectiveness varies on some tasks on each model.} To evaluate the effectiveness of the iterative process, we compared iterative pruning with direct pruning with recovery.  Direct pruning involves consecutively pruning all layers until the desired compression level is reached before the recovery phase, rather than iteratively pruning and recovering.

Table~\ref{tab:direct_onbyone} shows that iterative pruning generally preserves performance better than direct pruning on most language and reasoning tasks, such as \texttt{arc-easy} and \texttt{blimp}, while also maintaining \texttt{wikitext} perplexity even with increased layer dropping.  However, \texttt{winogrande} and \texttt{mmlu} do not benefit from iterative pruning, showing comparable or slightly reduced performance.  \texttt{arc-challenge} and \texttt{race} show comparable performance between the two methods. We hypothesize that our experimental setup, specifically training on \texttt{wikitext}, may lead to fitting on narrow knowledge, which is less suitable for these particular tasks.  Furthermore, \texttt{boolq} exhibits different behavior between Llama and Qwen, likely due to their distinct pre-training configurations. This suggests that the effectiveness of \algoname{} varies across tasks.

\section{\algoname{} Preservation Analysis} 

\begin{figure*} 
    \centering
    \begin{subfigure}{0.45\textwidth} 
        \centering
        \includegraphics[width=\textwidth]{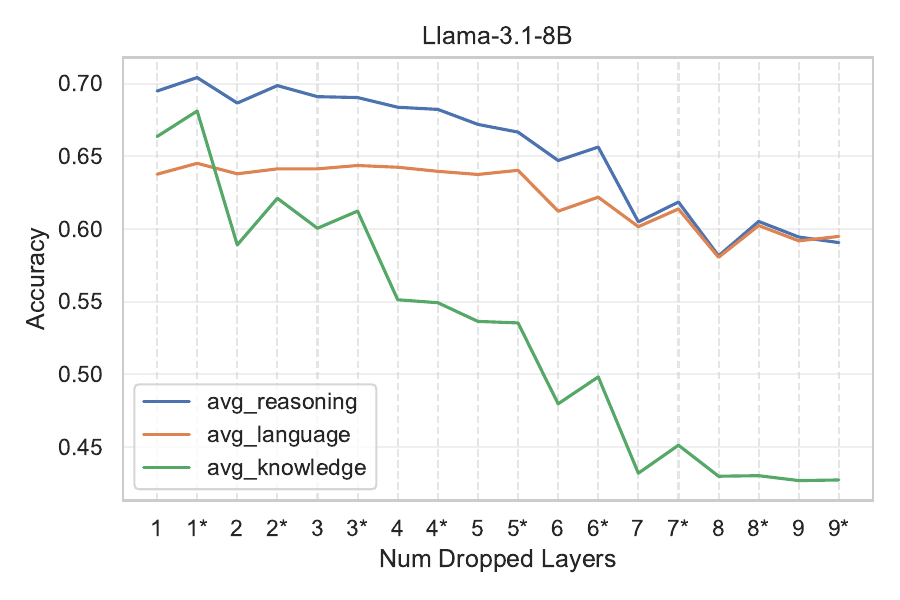}
    \end{subfigure}
    \hfill
    \begin{subfigure}{0.45\textwidth}
        \centering
        \includegraphics[width=\textwidth]{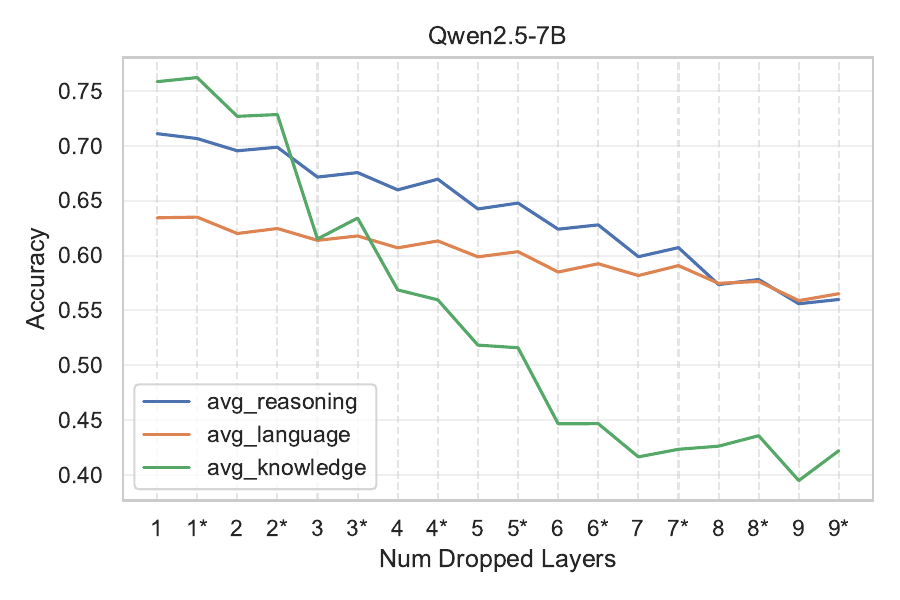}
    \end{subfigure}
    \caption{The average performance across pruning and recovery phase for 10 iterations on Llama 3.1-8B and Qwen2.5-7B on an average aggregation of reasoning, language, and knowledge tasks.}
    \label{fig:avg_trend}
\end{figure*}

\paragraph{\algoname{} effectively preserves language and reasoning abilities across iterations, though knowledge retention presents a challenge.} Figure~\ref{fig:avg_trend} shows the average performance trend across iterations for each task category.  While Qwen2.5-7B exhibits a slight, steady decrease (averaging \textasciitilde1\% per iteration) in reasoning and language task performance, Llama3.1-8B plateaus in language but shows a steady decline in reasoning.  Both models experience sharp performance drops in specific iterations (e.g., $M_{cs}^{[2]}$ for Llama and $M_{cs}^{[3]}$ for Qwen).  This affirms \algoname{}'s effectiveness in preserving language and reasoning abilities, though it suggests challenges in maintaining knowledge-based performance across iterations.

\paragraph{The recovery phase generally improves performance, though its impact is task and model dependent.}  The recovery phase generally improves performance by approximately 1\% for both models (Figure~\ref{fig:avg_trend}). However, its impact varies; for example, $M_{cs-rec}^{[5]}$ on Llama3.1-8B shows a slight decrease in reasoning performance after recovery, while language task performance increases. This indicates that the recovery process's effectiveness depends on the model family and the specific task.

\paragraph{\algoname{} Preserves and May Improves Linguistic Capabilites}  We evaluated the preservation of linguistic capacity across iterations using BLIMP, a benchmark consisting of 67 fine-grained linguistic problems.  We tested on Llama-3.1-8B and Qwen2.5-7B, categorizing the BLIMP subtasks into 13 groups for clearer visualization (see Appendix~\ref{apx:per-drop-pattern} for the groupings).

Overall, both models maintain or even improve scores across most categories in later iterations, surpassing the performance of the non-compressed models.  Furthermore, \algoname{} with recovery consistently outperforms the pruned model without recovery, with the exception of the "binding theory" category.  In this category, we observe a slight performance decay (\textasciitilde2\%) starting from the seventh iteration for Llama3.1-8B and the eighth iteration for Qwen2.5-7B.  The "coordinate structure" and "wh-that" categories exhibit differing trends between these family models. Llama3.1-8B shows an opposing trend at iteration 7 and beyond, with one subcategory plateauing while the other increases in performance.

\begin{figure*}
    \centering

    \begin{subfigure}{0.48\textwidth}
        \centering
        \includegraphics[width=\textwidth]{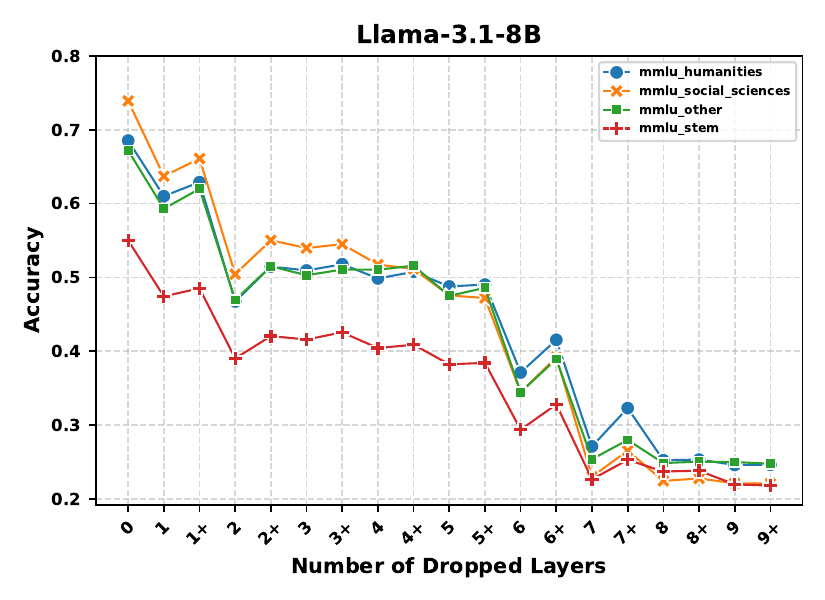}
    \end{subfigure}
    \hfill
    \begin{subfigure}{0.48\textwidth}
        \centering
        \includegraphics[width=\textwidth]{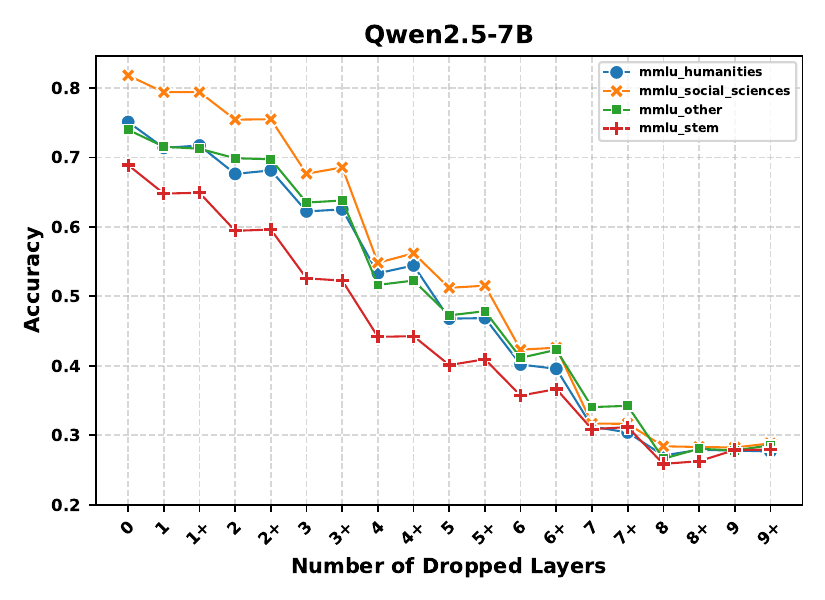}
    \end{subfigure}
    \vspace{-1em}
    \caption{Line charts depicts MMLU groupings performance on Llama-3.1-8B and Qwen2.5-7B in 10 iterations. "+" markers indicate the recovery phase; all other markers represent the pruning phase.}
    \label{fig:mmlu}
\end{figure*}

\paragraph{MMLU performance is sensitive to pruning, with recovery offering moderate gains across MMLU task categories} Figure~\ref{fig:mmlu} provides the MMLU performance across MMLU groupings. \footnote{using groupings defined in \texttt{lm-eval-harness}}  It shows that the pruning phase induces significant performance drops in some cases, notably in the early layer dropping of Llama3.1-8B (around 10\%) and from the third layer onward in Qwen2.5-7B. This suggests greater sensitivity of knowledge-based tasks to pruning. The subsequent recovery phase provides moderate improvements (about 2-3\%) for both models. Interestingly, Llama3.1-8B at $M_{cs-rec}^2$ shows a moderate performance gain, sustained across the next four iterations. This sustained improvement is not exhibited in Qwen2.5-7B, which instead exhibits a steady performance decline.  Performance trends across iterations are similar across MMLU categories within the same model, yet differ between models. These differences highlight model-specific variations in knowledge retention, potentially due to the distinct pre-training strategies of Llama3.1-8B and Qwen2.5-7B.

\begin{figure*}
    \centering
    \begin{subfigure}{0.49\textwidth}
        \centering
        \includegraphics[width=\textwidth]{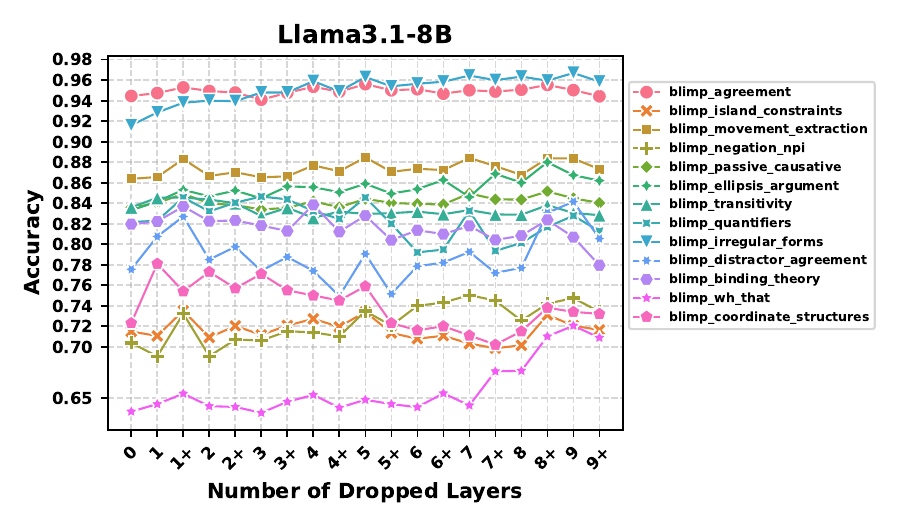}
    \end{subfigure}
    \hfill
    \begin{subfigure}{0.49\textwidth}
        \centering
        \includegraphics[width=\textwidth]{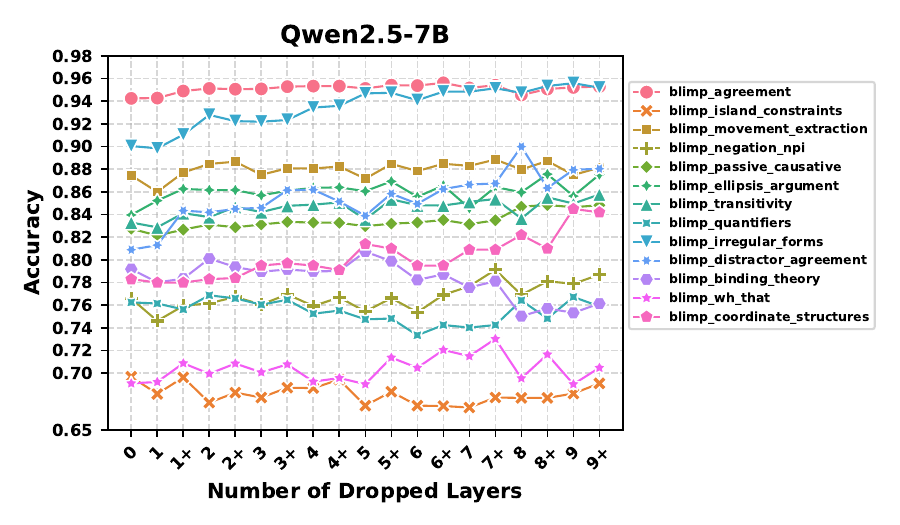}
    \end{subfigure}
    \vspace{-1em}
    \caption{Line charts showing BLIMP performance across 13 groupings for Llama-3.1-8B and Qwen-2.5-7B over 10 iterations. "+" markers indicate the recovery phase; all other markers represent the pruning phase.}
    \label{fig:blimp}
\end{figure*}

\paragraph{\algoname{} recovery improves multilingual performance, but the effect of improvement varies significantly across languages and tasks.} Given that our recovery phase uses English data, and both models possess multilingual capabilities, we investigated how much multilingual capacity is retained and whether \algoname{} induces zero-shot cross-lingual generalization during recovery. We evaluated our approach on three multilingual benchmarks: \texttt{XWinograd}, \texttt{XStoryCloze}, and \texttt{XNLI}. 

\begin{table}
  \centering
  \footnotesize
  \label{tab:results}
  \begin{tabular}{@{}>{\raggedright}p{1.7cm}|l|c|S[table-format=2.2]|S[table-format=2.2]|S[table-format=2.2]@{}}
    \toprule
    \textbf{Model} & \textbf{Approach} & \textbf{L} & \textbf{xW} & \textbf{xSc} & \textbf{xNLI} \\
    \midrule
    Llama3.1-8B 
    & Non-pruned & 32 & 81.43 & 63.61 & 45.65 \\
    \cmidrule{2-6}
    & LaCO & 24 & 67.39 & 52.05 & 37.78 \\
    & S-GPT & 24 & 56.37 & 48.80 & 34.25 \\
    & Ours P & 24 & 66.40 & 51.65 & 37.45 \\
    & Ours P+R & 24 & \textbf{71.68} & \textbf{55.53} & \textbf{39.77} \\
    \addlinespace[0.5ex]
    \midrule
    Qwen2.5-7B
    & Non-pruned & 28 & 81.48 & 62.04 & 43.37 \\
    \cmidrule{2-6}
    & LaCO & 22 & 64.71 & 51.66 & 36.49 \\
    & S-GPT & 21 & 66.33 & 55.28 & 37.32 \\
    & Ours-P & 21 & 65.54 & 53.48 & 36.99 \\
    & Ours-P+R & 21 & \textbf{72.26} & \textbf{55.76} & \textbf{39.46} \\
    \bottomrule
    \end{tabular}
    \caption{Performance Comparison in Multilingual Data. XW denotes \texttt{XWinograd} and XSc denotes \texttt{XStoryCloze}. Ours denotes \algoname{}, with \textbf{R} and \textbf{P} denotes running it with recovery and pruning phases, respectively.  }
    \label{tab:multilingual_all}
\end{table}

Table~\ref{tab:multilingual_all} compares the baseline models with \algoname{} (without recovery). The results demonstrate that our recovery method improves performance by 5-6\% on \texttt{XWinograd} across both models, with 2-5\% improvements on \texttt{XStoryCloze} and \texttt{XNLI}. These findings suggest effective generalization to multilingual data from English-based recovery.

Figure~\ref{fig:radar_multilingual} presents performance per language within each dataset. Both Llama3.1-8B and Qwen2.5-7B show similar patterns, though Llama3.1-8B exhibits larger performance gaps. For Llama3.1-8B, English shows the strongest performance preservation, with approximately 8\% improvements on \texttt{XWinograd}, \texttt{XStoryCloze}, and \texttt{XNLI} compared to the non-recovery baseline. Other languages show more modest gains, averaging around 2\% post-recovery. Notably, some languages in Llama 3.1-8B's \texttt{XStoryCloze}, particularly \texttt{es} and \texttt{id}, show improvements comparable to \texttt{en} (6-8\%). In \texttt{xnli}, \texttt{de} and \texttt{en}, performances approach those of the unpruned models. These findings align with \citet{choenni-etal-2023-languages}, suggesting varying cross-lingual influence. Some languages, such as \texttt{hi} and \texttt{ur}, show minimal preservation (less than 1\% improvement). \texttt{zh} notably performs worse with recovery compared to layer pruning alone. These results indicate that while English-based recovery can facilitate zero-shot cross-lingual generalization, its effectiveness varies considerably across languages and tasks.

More analysis can be seen in Appendix~\ref{sec:apx-more-analysis}

\section{Related Works}

Model pruning has gained significant attention recently due to the emergence of Large Language Models (LLMs). One of the approaches is to do unit size reduction, where several approaches leverage dimensionality reduction techniques~\cite{lin2024modegptmodulardecompositionlarge, ashkboos2024slicegpt} to compress weight matrices, thereby reducing hidden unit dimensions. Various metrics have been explored to identify prunable weights, including Hessian information \cite{frantar2023sparsegptmassivelanguagemodels,ling2024slimgptlayerwisestructuredpruning}, Kronecker-factored curvature~\cite{vanderouderaa2024llmsurgeon}, and magnitude information~\cite{sun2024simpleeffectivepruningapproach, guo2024owllargelanguagemodel}. On the other hand,  block pruning is done by employing some metrics, such as Hessian information~\cite{ma2023llmprunerstructuralpruninglarge}, output similarity~\cite{yang-etal-2024-laco, men2024shortgptlayerslargelanguage}, and learnable parameters to determine block significance~\cite{liu2024foldgptsimpleeffectivelarge, xia2024shearedllamaacceleratinglanguage}. Some approaches opt to merge blocks instead of removing them~\cite{yang-etal-2024-laco,chen-etal-2024-lemon}. \citealp{muralidharan2024compactlanguagemodelspruning} combines iterative pruning with Neural Architecture Search~\cite{elsken2019neuralarchitecturesearchsurvey}, utilizing multiple metrics for model compression. Many of these techniques incorporate recovery phase~\cite{ling2024slimgptlayerwisestructuredpruning,sun2024simpleeffectivepruningapproach,yin2024outlierweighedlayerwisesparsity,ma2023llmprunerstructuralpruninglarge,muralidharan2024compactlanguagemodelspruning}. In our work, we adopt an iterative approach based on output similarity, followed by a recovery process, prioritizing simplicity and minimal computational requirements.

\section{Conclusion }
This work introduced \algoname{}, a simple yet effective iterative block pruning method designed for simple and efficient LLM compression. \algoname{} outperforms other baselines requiring only 2.5M tokens.   Furthermore, our analysis reveals distinct pruning patterns on the observed tasks across different model architectures and also preserves the multilingual capabilities, even with English-only recovery data.

\section*{Limitations}
We only observe the Qwen2.5 and Llama3 family models. Additionally, we only observe Wikitext, to perform the pruning phase (following \citealp{yang-etal-2024-laco}) and recovery phase. We leave the possibility of using a variety of dataset sizes or sampling techniques for this technique for future work.

\section*{Ethics Statement}
This work has no ethical issues, as we propose to perform a compression technique. The data used do not contain personally identifiable information or offensive content. The artifacts we utilize are consistent with intended use and adhere to the license usage (research purpose). We use AI Assistants (LLMs, Grammarly, and Writefull) to assist our writing in correcting grammatical errors.

\bibliography{anthology,custom}
\bibliographystyle{acl_natbib}

\appendix

\section{Calibration Dataset}\label{sec: apx-calibration}
Here is the calibration dataset sampled from Wikitext used to run the pruning phase. These are sampled uniformly.
\begin{enumerate}
    \item " = There 's Got to Be a Way = "
    \item " Cullen is the namesake of the John Cullen Award , previously given to key IHL players."
    \item "Ancient Egyptian deities are the gods and goddesses worshipped in ancient Egypt"
    \item "Nationally important deities gave rise to local manifestations"
    \item "The first aerodrome in the UK was established by the Aero Club at Muswell Manor on the Isle of Sheppey"
    \item "Competitive gliding in the UK takes place between May and September"
    \item "Most aerodromes used for public transport operations are required to be licensed by the CAA"
    \item "Within this framework certain sectors of GA are governed on a devolved basis"
    \item "James Pollock , in his final report as Mint Director in 1873"
    \item "Stability of the ylide with higher stability similarly leading to greater reversibility."
\end{enumerate}

\section{Performance Trend for Each Iteration} \label{apx:per-drop-pattern}

\begin{figure*}
    \centering

    \begin{subfigure}{\textwidth}
        \centering
        \includegraphics[width=0.95\textwidth]{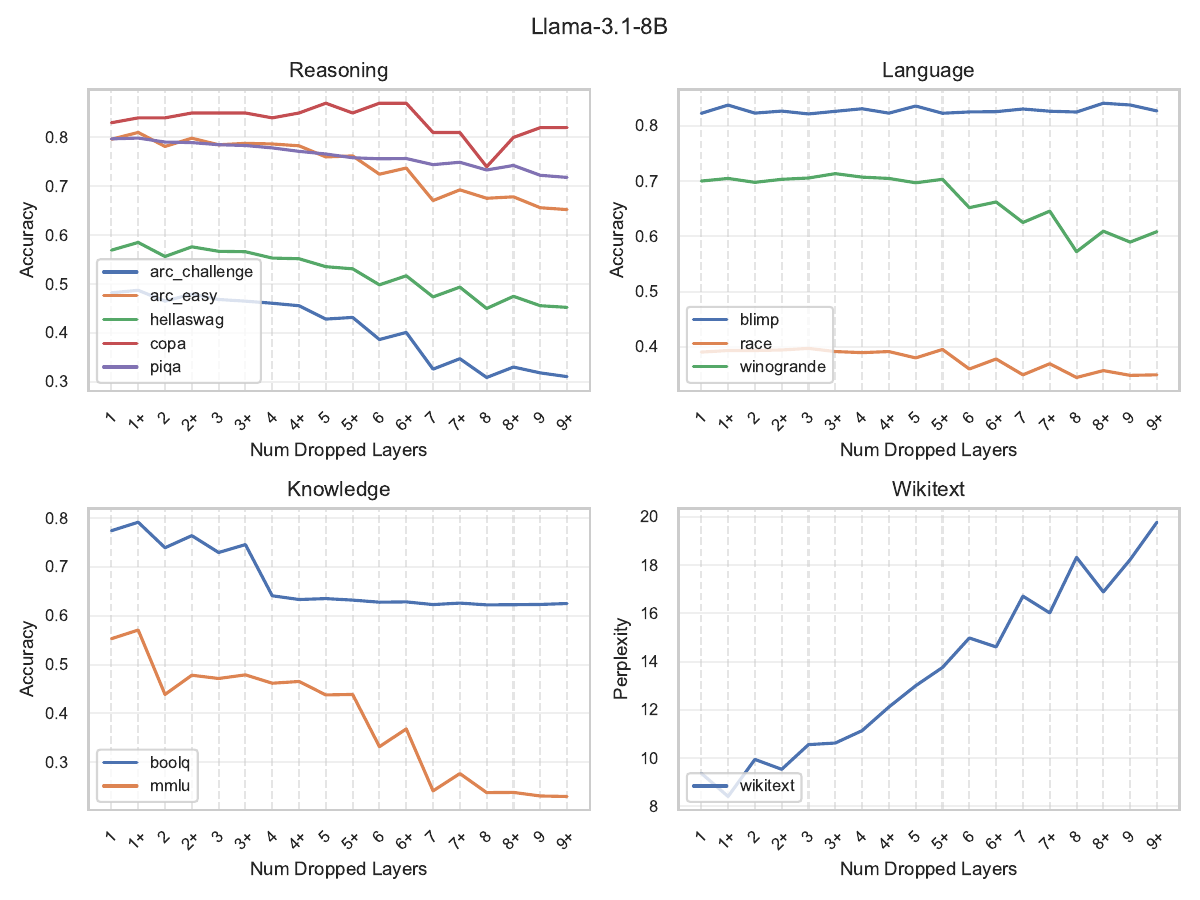}
    \end{subfigure}
    \hfill
    \begin{subfigure}{\textwidth}
        \centering
        \includegraphics[width=\textwidth]{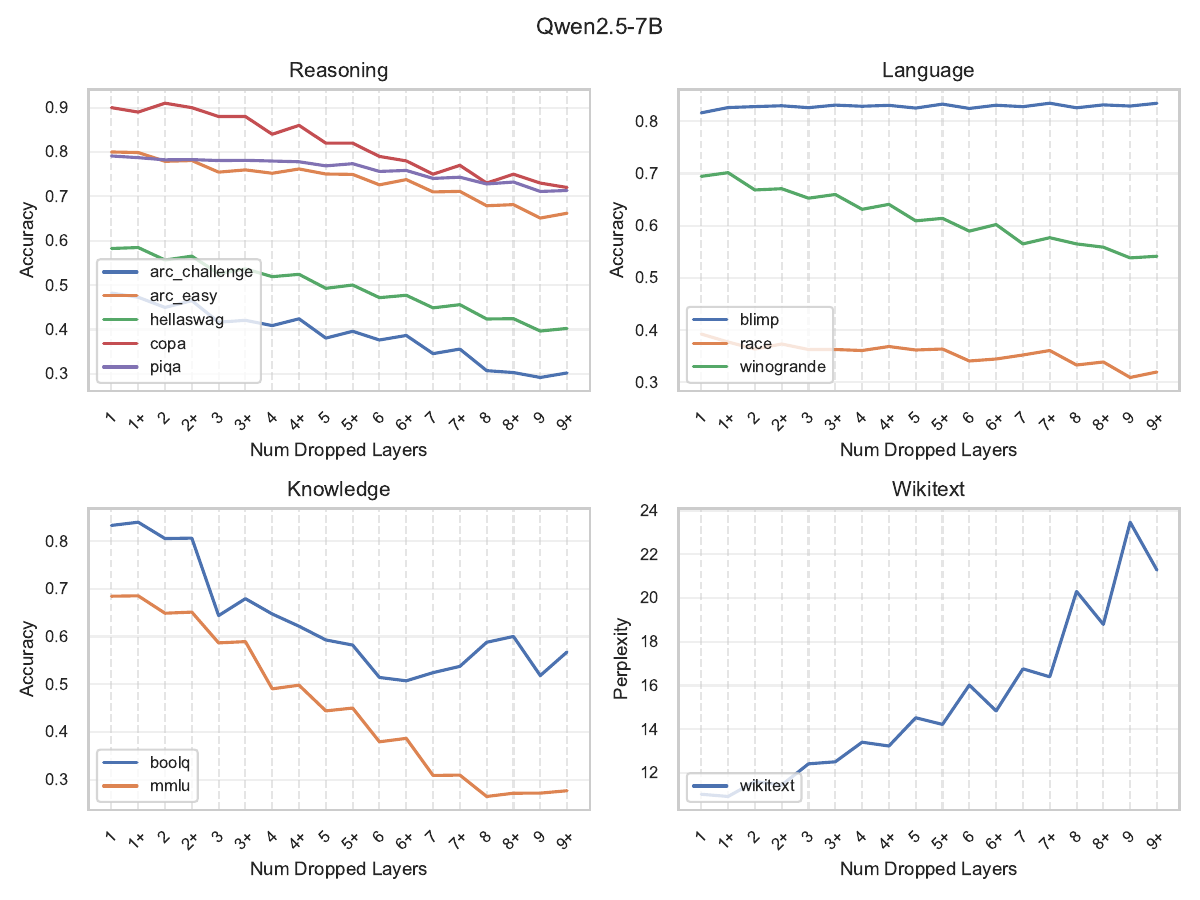}
    \end{subfigure}
    \caption{The performance across pruning and recovery phase for 10 iterations in Qwen2.5-7B and Llama3.1-8B.}
    \label{fig:full_trend}
\end{figure*}

The fine-grained performance trend can be seen in Figure~\ref{fig:full_trend}.

\section{Blimp Categories} \label{apx:blimp-cat}

Blimp categories that we define can be seen in Table~\ref{tab:blimp_agreement},\ref{tab:blimp_syntax},\ref{tab:blimp_argument},\ref{tab:blimp_special}.
\begin{table*}[htbp]
\centering
\small
\begin{tabular}{|p{0.25\textwidth}|p{0.65\textwidth}|}
\hline
\textbf{Group} & \textbf{Tests} \\
\hline
blimp\_agreement & 
\begin{itemize}[noitemsep]
    \item blimp\_regular\_plural\_subject\_verb\_agreement\_1
    \item blimp\_regular\_plural\_subject\_verb\_agreement\_2
    \item blimp\_irregular\_plural\_subject\_verb\_agreement\_1
    \item blimp\_irregular\_plural\_subject\_verb\_agreement\_2
    \item blimp\_determiner\_noun\_agreement\_1
    \item blimp\_determiner\_noun\_agreement\_2
    \item blimp\_determiner\_noun\_agreement\_irregular\_1
    \item blimp\_determiner\_noun\_agreement\_irregular\_2
    \item blimp\_determiner\_noun\_agreement\_with\_adj\_2
    \item blimp\_determiner\_noun\_agreement\_with\_adj\_irregular\_1
    \item blimp\_determiner\_noun\_agreement\_with\_adj\_irregular\_2
    \item blimp\_determiner\_noun\_agreement\_with\_adjective\_1
    \item blimp\_anaphor\_gender\_agreement
    \item blimp\_anaphor\_number\_agreement
\end{itemize} \\
\hline
blimp\_distractor\_agreement & 
\begin{itemize}[noitemsep]
    \item blimp\_distractor\_agreement\_relational\_noun
    \item blimp\_distractor\_agreement\_relative\_clause
\end{itemize} \\
\hline
\end{tabular}
\caption{BLiMP Agreement Tests}
\label{tab:blimp_agreement}
\end{table*}

\begin{table*}[htbp]
\centering
\small
\begin{tabular}{|p{0.25\textwidth}|p{0.65\textwidth}|}
\hline
\textbf{Group} & \textbf{Tests} \\
\hline
blimp\_island\_constraints & 
\begin{itemize}[noitemsep]
    \item blimp\_wh\_island
    \item blimp\_complex\_NP\_island
    \item blimp\_adjunct\_island
    \item blimp\_sentential\_subject\_island
    \item blimp\_left\_branch\_island\_echo\_question
    \item blimp\_left\_branch\_island\_simple\_question
\end{itemize} \\
\hline
blimp\_movement\_extraction & 
\begin{itemize}[noitemsep]
    \item blimp\_wh\_questions\_object\_gap
    \item blimp\_wh\_questions\_subject\_gap
    \item blimp\_wh\_questions\_subject\_gap\_long\_distance
    \item blimp\_coordinate\_structure\_constraint\_object\_extraction
    \item blimp\_existential\_there\_subject\_raising
    \item blimp\_existential\_there\_object\_raising
    \item blimp\_expletive\_it\_object\_raising
\end{itemize} \\
\hline
blimp\_wh\_that & 
\begin{itemize}[noitemsep]
    \item blimp\_wh\_vs\_that\_no\_gap
    \item blimp\_wh\_vs\_that\_no\_gap\_long\_distance
    \item blimp\_wh\_vs\_that\_with\_gap
    \item blimp\_wh\_vs\_that\_with\_gap\_long\_distance
\end{itemize} \\
\hline
\end{tabular}
\caption{BLiMP Syntax and Movement Tests}
\label{tab:blimp_syntax}
\end{table*}

\begin{table*}[htbp]
\centering
\small
\begin{tabular}{|p{0.25\textwidth}|p{0.65\textwidth}|}
\hline
\textbf{Group} & \textbf{Tests} \\
\hline
blimp\_passive\_causative & 
\begin{itemize}[noitemsep]
    \item blimp\_passive\_1
    \item blimp\_passive\_2
    \item blimp\_animate\_subject\_passive
    \item blimp\_causative
\end{itemize} \\
\hline
blimp\_transitivity & 
\begin{itemize}[noitemsep]
    \item blimp\_transitive
    \item blimp\_intransitive
    \item blimp\_inchoative
    \item blimp\_animate\_subject\_trans
\end{itemize} \\
\hline
blimp\_irregular\_forms & 
\begin{itemize}[noitemsep]
    \item blimp\_irregular\_past\_participle\_adjectives
    \item blimp\_irregular\_past\_participle\_verbs
\end{itemize} \\
\hline
\end{tabular}
\caption{BLiMP Argument Structure and Form Tests}
\label{tab:blimp_argument}
\end{table*}

\begin{table*}[htbp]
\centering
\small
\begin{tabular}{|p{0.25\textwidth}|p{0.65\textwidth}|}
\hline
\textbf{Group} & \textbf{Tests} \\
\hline
blimp\_negation\_npi & 
\begin{itemize}[noitemsep]
    \item blimp\_npi\_present\_1
    \item blimp\_npi\_present\_2
    \item blimp\_only\_npi\_licensor\_present
    \item blimp\_only\_npi\_scope
    \item blimp\_sentential\_negation\_npi\_licensor\_present
    \item blimp\_sentential\_negation\_npi\_scope
    \item blimp\_matrix\_question\_npi\_licensor\_present
\end{itemize} \\
\hline
blimp\_quantifiers & 
\begin{itemize}[noitemsep]
    \item blimp\_superlative\_quantifiers\_1
    \item blimp\_superlative\_quantifiers\_2
    \item blimp\_existential\_there\_quantifiers\_1
    \item blimp\_existential\_there\_quantifiers\_2
\end{itemize} \\
\hline
blimp\_binding\_theory & 
\begin{itemize}[noitemsep]
    \item blimp\_principle\_A\_c\_command
    \item blimp\_principle\_A\_case\_1
    \item blimp\_principle\_A\_case\_2
    \item blimp\_principle\_A\_domain\_1
    \item blimp\_principle\_A\_domain\_2
    \item blimp\_principle\_A\_domain\_3
    \item blimp\_principle\_A\_reconstruction
\end{itemize} \\
\hline
blimp\_ellipsis\_argument & 
\begin{itemize}[noitemsep]
    \item blimp\_ellipsis\_n\_bar\_1
    \item blimp\_ellipsis\_n\_bar\_2
    \item blimp\_drop\_argument
\end{itemize} \\
\hline
blimp\_coordinate\_structures & 
\begin{itemize}[noitemsep]
    \item blimp\_coordinate\_structure\_constraint\_complex\_left\_branch
\end{itemize} \\
\hline
\end{tabular}
\caption{BLiMP Specialized Construction Tests}
\label{tab:blimp_special}
\end{table*}

\section{Granular Multilingual Results}

\begin{figure*}
    \centering

    \begin{subfigure}{0.32\textwidth}
        \centering
        \includegraphics[width=\textwidth]{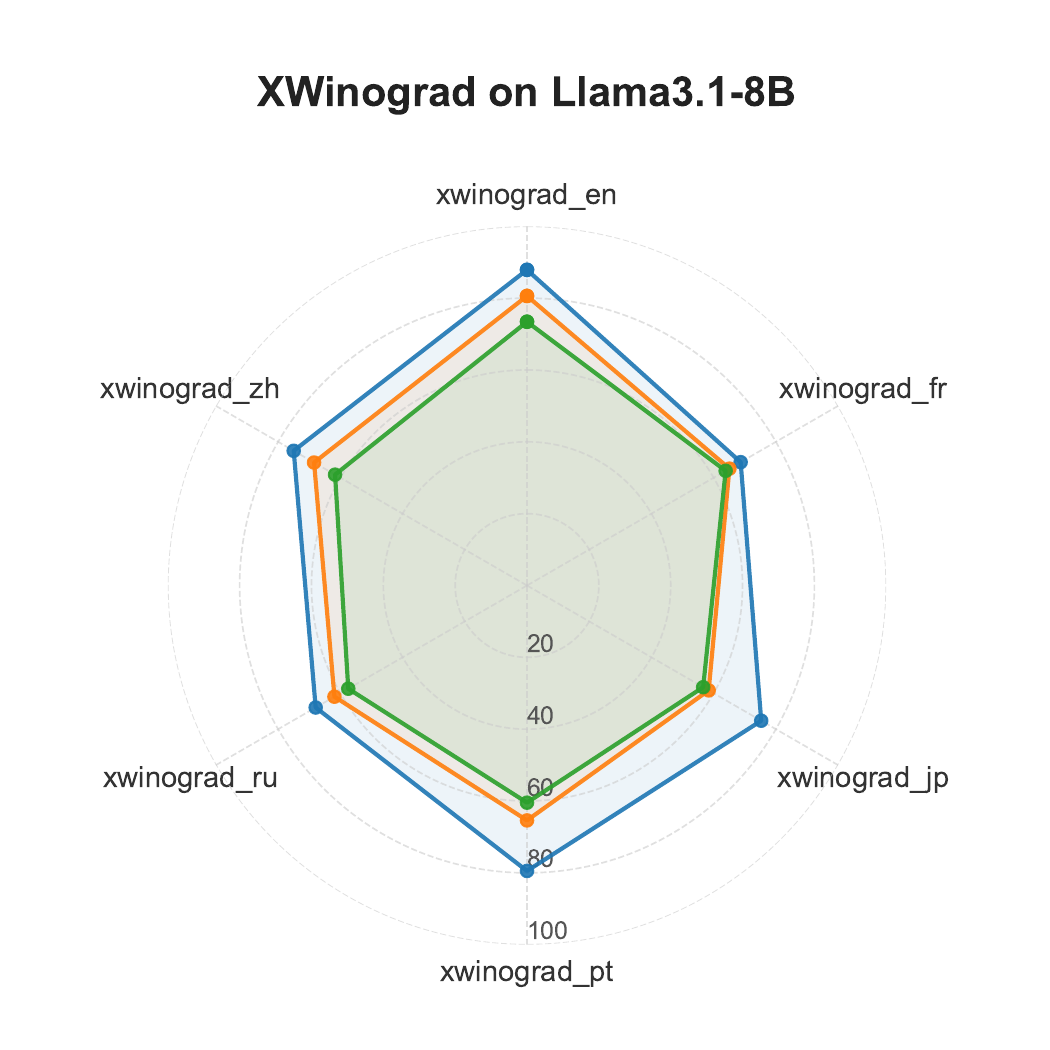}
    \end{subfigure}
    \hfill
    \begin{subfigure}{0.32\textwidth}
        \centering
        \includegraphics[width=\textwidth]{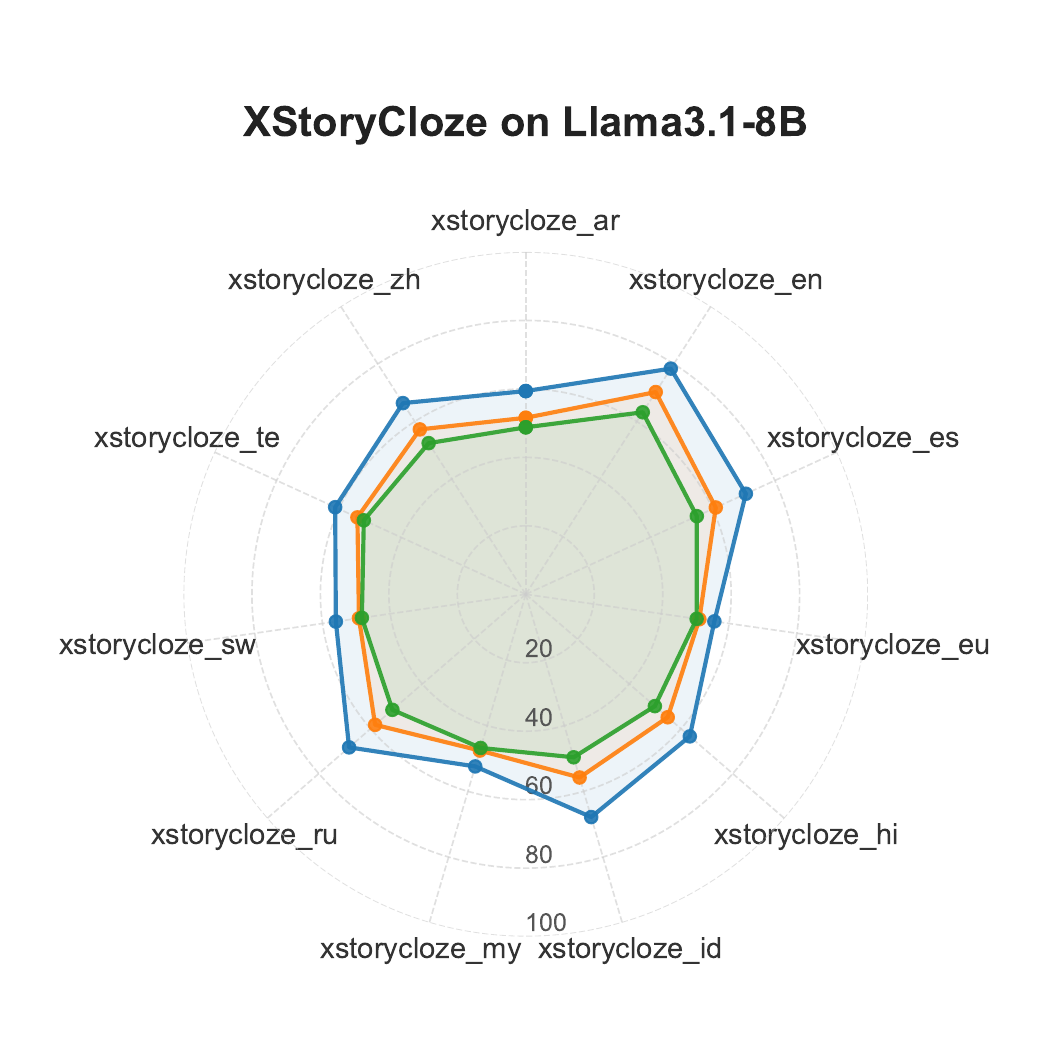}
    \end{subfigure}
    \begin{subfigure}{0.32\textwidth}
        \centering
        \includegraphics[width=\textwidth]{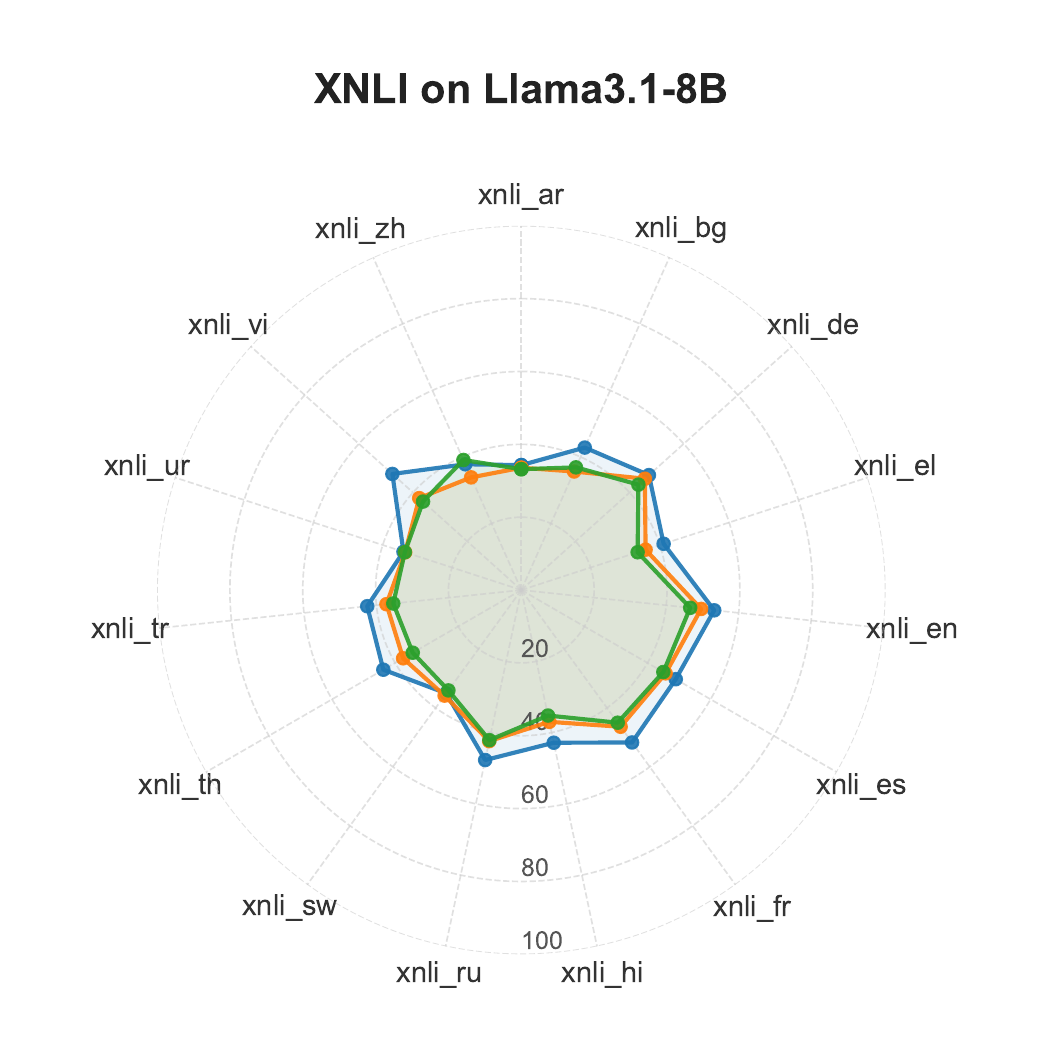}
    \end{subfigure}
    \vspace{-1em}
    
    \begin{subfigure}{0.32\textwidth}
        \centering
        \includegraphics[width=\textwidth]{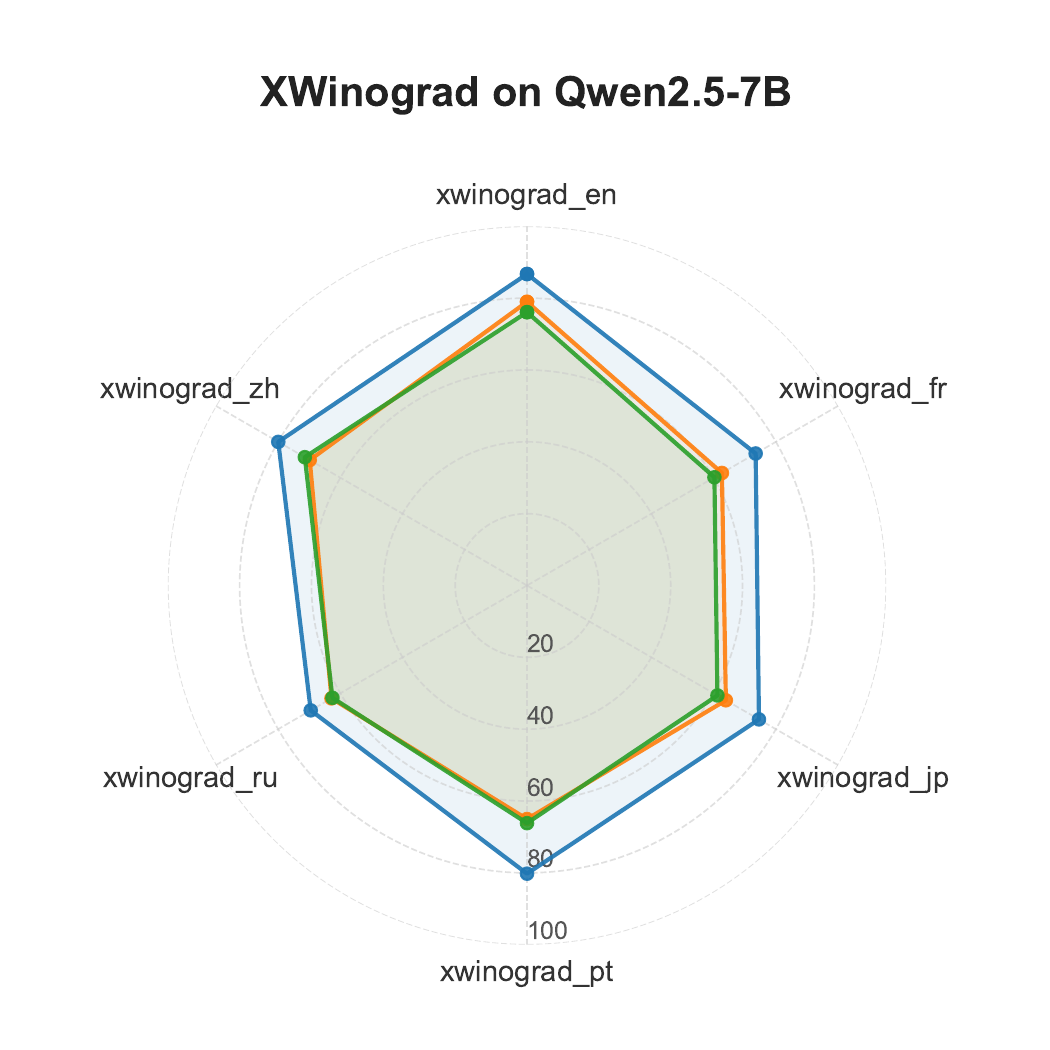}
    \end{subfigure}
    \hfill
    \begin{subfigure}{0.32\textwidth}
        \centering
        \includegraphics[width=\textwidth]{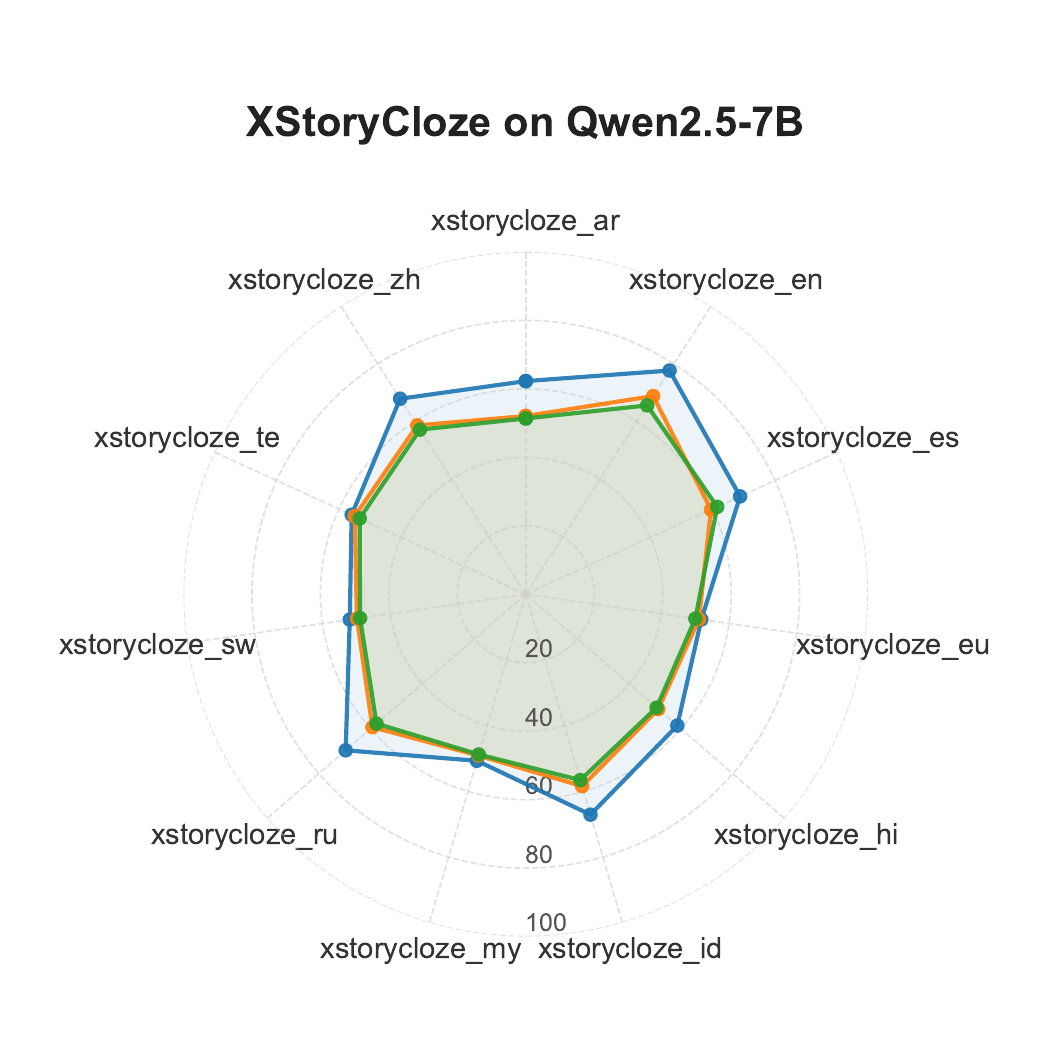}
    \end{subfigure}
    \begin{subfigure}{0.32\textwidth}
        \centering
        \includegraphics[width=\textwidth]{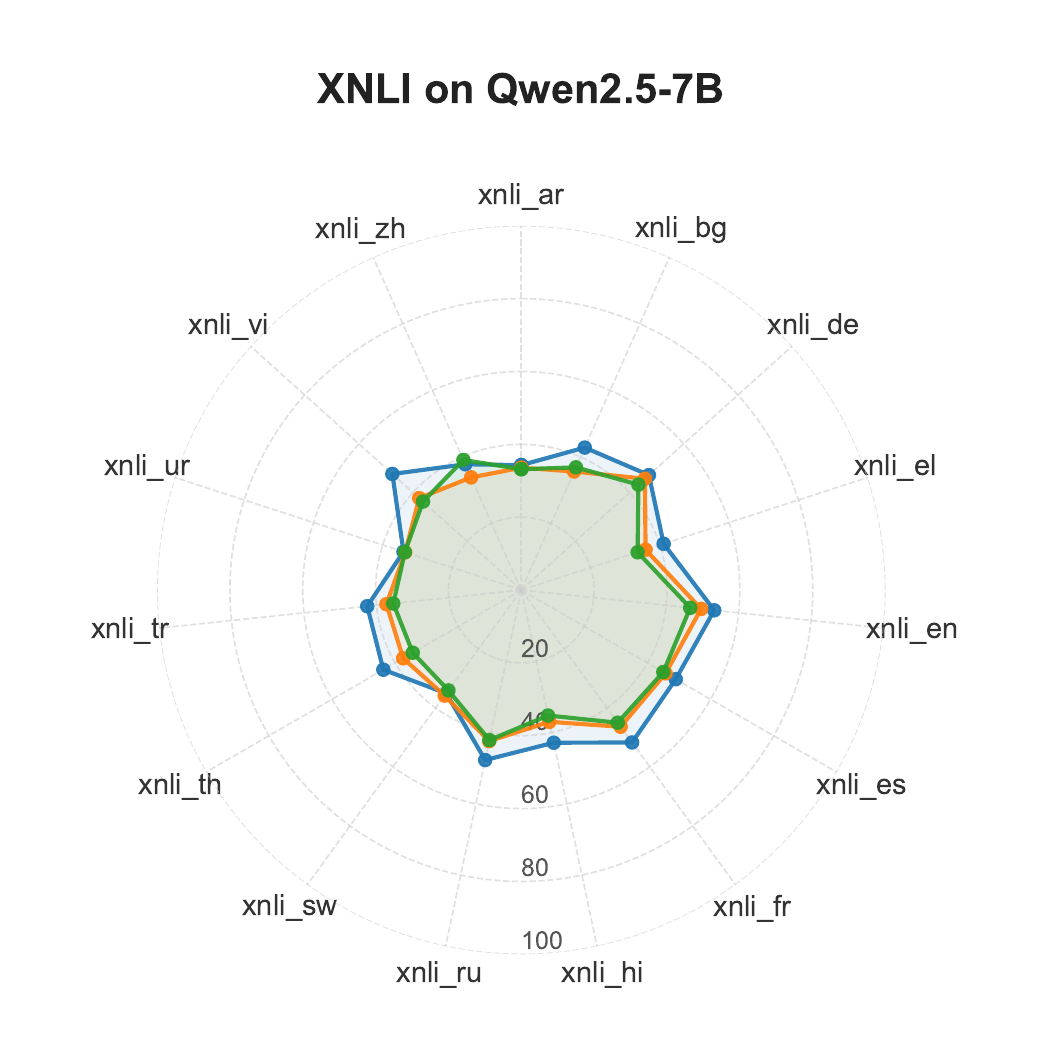}
    \end{subfigure}
    \vspace{-1em}
    \caption{Granular multilingual performance (Accuracy) across XWinograd, XStoryCloze and XNLI on Qwen2.5-7B. \textcolor{blue-multi}, \textcolor{brown-multi}, and \textcolor{green-multi} denotes result of non-pruned model, \algoname{} with recovery phase, and \algoname{} without recovery phase, respectively.}
    \label{fig:radar_multilingual}
\end{figure*}

The Granular multilingual across \texttt{xwinograd}, \texttt{xstorycloze}, \texttt{xnli} analysis can be viewed in Figure~\ref{fig:radar_multilingual}

\section{Layer Mapping in Recovery Phase}\label{sec:apx-mapping}
To define the mapping function \( \text{map}(l) \) in iteration \( j \), we aim to align the student's layer index \( l \) with the corresponding original index in the teacher model. However, if any layers in the teacher model with indices lower than \( l \) were dropped before iteration \( j \), the mapping must account for these dropped layers. Specifically, \( \text{map}(l) \) is adjusted by increasing it by the number of dropped layers with indices less than \( \text{map}(l) \). For example, if the dropped layer indices are \( [3, 4] \) and \( l = 10 \), then \( \text{map}(10) = 12 \), as the two dropped layers shift the mapping while \( map(1) = 1 \). Formally, let \( D \) be the set of dropped layer indices in the teacher model before iteration \( j \), sorted in ascending order. The function \( \text{map}(l) \) maps the student's layer index \( l \) to the teacher's original index \( m \), where \( m \) is the unique solution to the equation \( m = l + \left| \{ d \in D \mid d < m \} \right| \). 

\section{More Analysis} \label{sec:apx-more-analysis}

\begin{figure*}
    \centering
    \begin{subfigure}{0.49\textwidth}
        \centering
        \includegraphics[width=\textwidth]{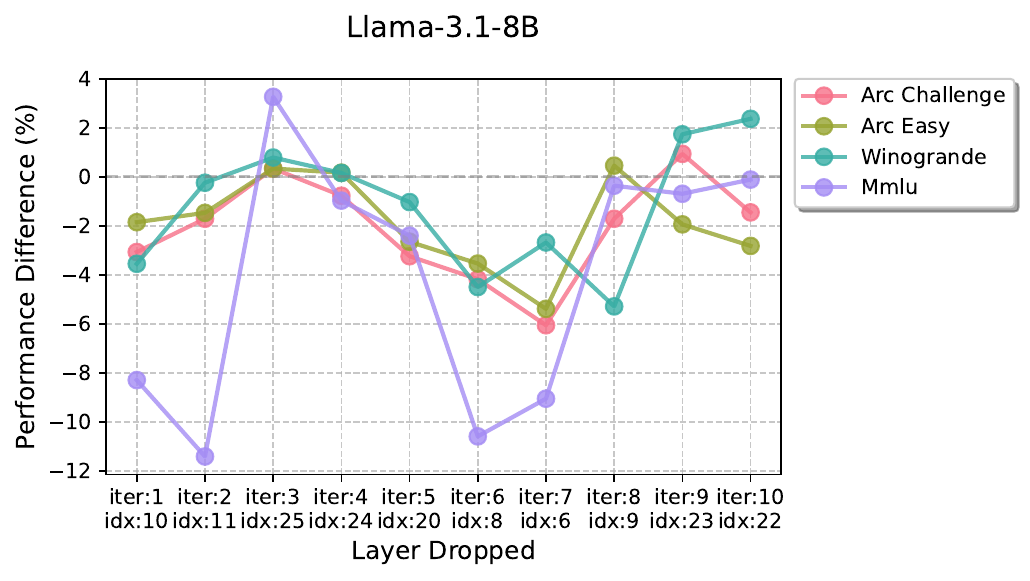}
    \end{subfigure}
    \begin{subfigure}{0.49\textwidth}
        \centering
        \includegraphics[width=\textwidth]{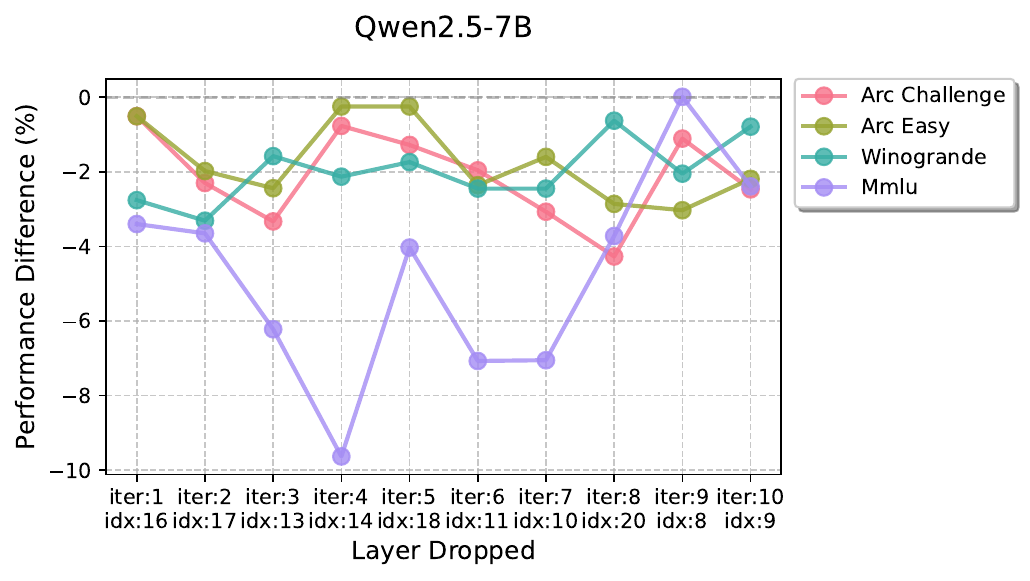}
    \end{subfigure}
    \caption{The performance differences between before and after two phases done for each iteration (\texttt{iter}) in \algoname{} on LLAMA 3-1-8B and Qwen 2.5-7B. idx denoted the index of the dropped layer (starts from 0).}
    \label{fig: delta}
\end{figure*}
\paragraph{\algoname{} exhibits task-specific layer sensitivities that vary between models.} We investigated which layer drops correlate with significant performance declines, indicating layer importance. Figure~\ref{fig: delta} shows performance differences across tasks and categories for Qwen2.5-7B and Llama-3.1-8B, revealing distinct drop patterns for each model. Llama3.1-8B's performance drops tend to occur in the lower half of its layers, while Qwen's are concentrated in the upper half.  Specifically, Llama3.1-8B shows significant drops on \texttt{arc-easy} and \texttt{arc-challenge} in iteration 1, 6, and 7, and on \texttt{winogrande} in iteration 1, 6, and 8.  MMLU on Llama3.1-8B shows steep declines in iteration 10 and 11 during early iterations, followed by improvement and stagnation.  Qwen2.5-7B exhibits different trends, with notable (>5\%) decreases on MMLU in iteration 3, 4, 6, and 7.

\end{document}